\documentclass[letterpaper]{article} 
\usepackage{aaai2026}  
\usepackage{times}  
\usepackage{helvet}  
\usepackage{courier}  
\usepackage[hyphens]{url}  
\usepackage{graphicx} 
\usepackage{natbib}  
\usepackage{caption} 
\usepackage{algorithm}
\usepackage{algorithmic}
\usepackage{amsmath,amssymb,mathtools,amsthm}
\theoremstyle{plain}
\newtheorem{theorem}{Theorem}

\newtheorem{corollary}[theorem]{Corollary}
\theoremstyle{definition}
\newtheorem{definition}[theorem]{Definition}

\theoremstyle{remark}

\usepackage{multirow}
\usepackage{arydshln}
\usepackage{colortbl}
\definecolor{Gray}{gray}{0.95}
\newcommand{\gb}[1]{{\cellcolor{gray!18}{#1}}}
\usepackage{booktabs}
\usepackage[table]{xcolor}

\usepackage{newfloat}
\usepackage{listings}
\DeclareCaptionStyle{ruled}{labelfont=normalfont,labelsep=colon,strut=off} 
\floatstyle{ruled}
\newfloat{listing}{tb}{lst}{}
\floatname{listing}{Listing}
\title{RelaCtrl: Relevance-Guided Efficient Control for Diffusion Transformers}
\author{
Ke Cao\textsuperscript{1,2,3*}, Jing Wang\textsuperscript{3*}, Ao Ma\textsuperscript{3*\dag}, Jiasong Feng\textsuperscript{3}, Xuanhua He\textsuperscript{1}, \\ Run Ling\textsuperscript{3}, Haowei Liu\textsuperscript{3}, Jian Lu\textsuperscript{3}, Wei Feng\textsuperscript{4}, Haozhe Wang\textsuperscript{4}, Hongjuan Pei\textsuperscript{4}, Yihua Shao\textsuperscript{3}, \\ Zhanjie Zhang\textsuperscript{3,5\ddag}, Jie Zhang\textsuperscript{1,2,6\ddag}}

\affiliations{
        \textsuperscript{\rm 1}University of Science and Technology of
China \\
        \textsuperscript{\rm 2}Hefei Institutes of Physical Science, Chinese Academy of Sciences \\
        \textsuperscript{\rm 3}360 AI Research \\
        \textsuperscript{\rm 4}University of Chinese Academy of Sciences \\
        \textsuperscript{\rm 5}Zhejiang University\\
        \textsuperscript{\rm 6}Zhongke Hefei Institute of Technology Innovation Engineering \\
    
    \{caoke200820, zhangzhanj\}@126.com, zhangjie@iim.ac.cn
}

\usepackage{bibentry}

\begin{document}

\maketitle

\begingroup
\renewcommand\thefootnote{}%
\footnotetext{%
\textsuperscript{*} Equal contribution.\par
\hspace{\parindent}\textsuperscript{\dag} Project leader.\par
\hspace{\parindent}\textsuperscript{\ddag} Corresponding author.%
}
\endgroup
\begin{abstract}
The Diffusion Transformer plays a pivotal role in advancing text-to-image and text-to-video generation, owing primarily to its inherent scalability. However, existing controlled diffusion transformer methods incur significant parameter and computational overheads and suffer from inefficient resource allocation due to their failure to account for the varying relevance of control information across different transformer layers.
To address this, we propose the Relevance-Guided Efficient Controllable Generation framework, \textbf{RelaCtrl}, enabling efficient and resource-optimized integration of control signals into the Diffusion Transformer.
First, we evaluate the relevance of each layer in the Diffusion Transformer to the control information by assessing the ControlNet Relevance Score, which measures the impact of skipping each control layer on both the quality of generation and the control effectiveness during inference.
Based on the strength of the relevance, we then tailor the positioning, parameter scale, and modeling capacity of the control layers to reduce unnecessary parameters and redundant computations. Additionally, to further improve efficiency, we replace the self-attention and FFN in the commonly used copy block with the carefully designed Two-Dimensional Shuffle Mixer (\textbf{TDSM}), enabling efficient implementation of both the token mixer and channel mixer. Both qualitative and quantitative experimental results demonstrate that our approach achieves superior performance with only 15\% of the parameters and computational complexity compared to PixArt-$\delta$. \end{abstract}

\section{Introduction}
The Diffusion Transformer (DiT)~\cite{peebles2023dit}, with its strong scalability and multi-modal alignment capabilities, has significantly advanced the fields of text-to-image and text-to-video generation (like, PixArt-$\alpha$~\cite{chen2023PixArta}, Flux~\cite{blackforestlabs2024flux}, Stable Diffusion 3~\cite{stabilityai2024sd3}, CogVideoX~\cite{yang2024cogvideox}, Sora~\cite{videoworldsimulators2024}, HunyuanVideo~\cite{li2024hunyuan}, and Qihoo-T2X~\cite{wang2024qihoo}, etc). By leveraging its robust architecture and scalability, the fidelity of the generated results and consistency with the given textual description are dramatically improved. Recent studies, such as PixArt-$\delta$~\cite{chen2024PixArtc} and OminiControl~\cite{tan2024ominicontrol}, focus on controlled text-to-image generation based on the DiT framework, promoting its application in real-world scenarios such as AI-driven content creation~\cite{ma2025lay2story, he2025plangen, wang2025learning} and e-commerce shopping~\cite{lu2025uni,bi2025customttt, zhang2025u}.
\begin{figure}[t]
\centering
\includegraphics[width=0.9\linewidth]{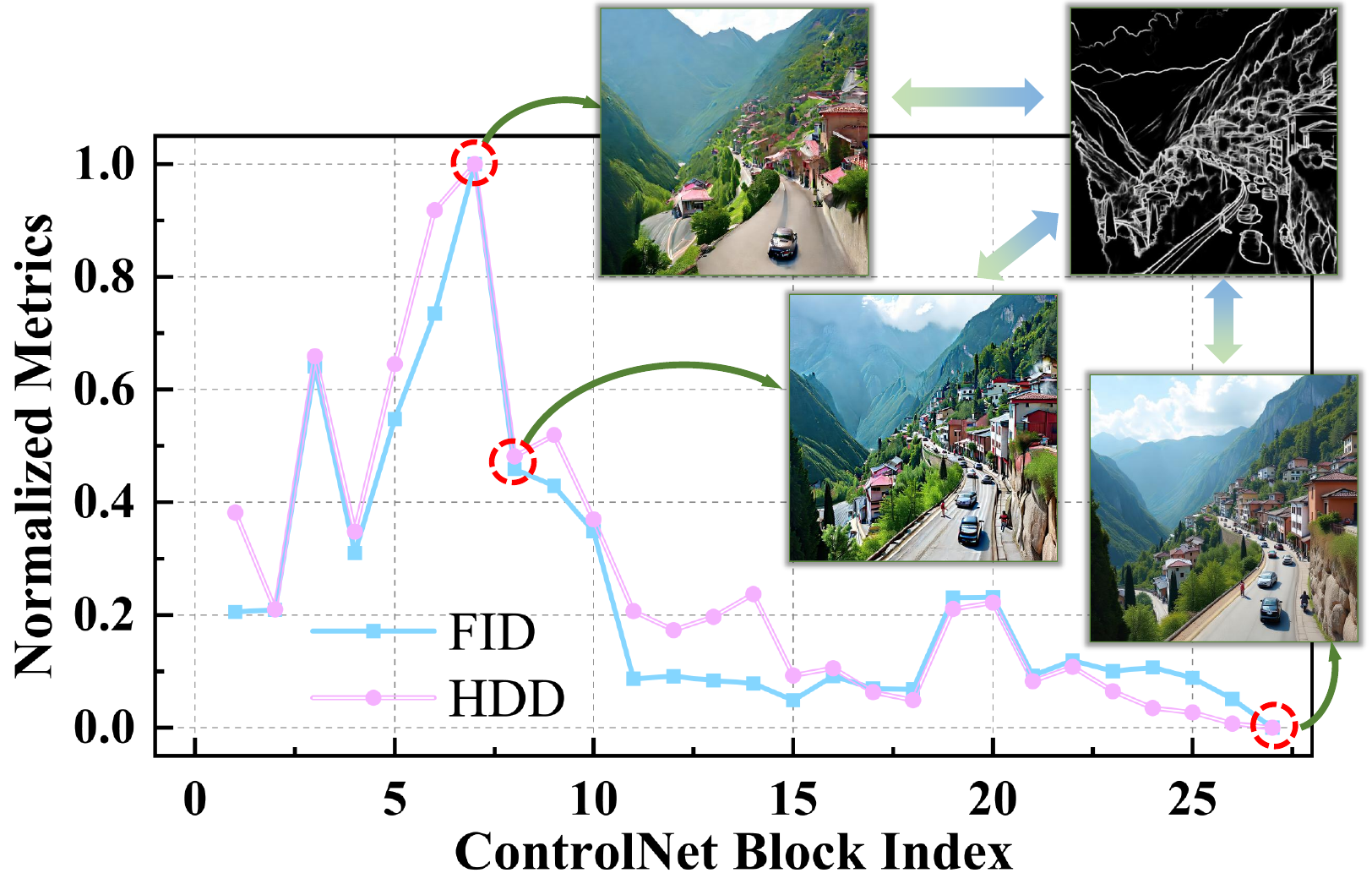}
\caption{\label{img:intro_fidandhdd}Effect of skipping a specific position within the ControlNet on the quality of the generated image. Higher FID and HDD indicate a more significant impact of the skipped layer on the quality of the final results, reflecting a stronger correlation with the generation.
}
\end{figure}

However, current controlled generation methods for DiT face two main shortcomings.
\textbf{Firstly}, a significant number of additional parameters and computations are introduced, increasing the burden on training and inference. For example, PixArt-$\delta$ directly duplicates the first half of the network's blocks (i.e., 13 blocks), resulting in a 50\% increase in both the number of parameters and computational complexity. Similarly, the control token concatenation in OminiControl adds only a limited number of parameters but doubles the number of tokens involved in the attention and linear layers, leading to a nearly 70\% increase in overall computational complexity.
\textbf{Secondly}, the varying relevance of control information across different layers of the network is often overlooked, resulting in inefficient allocation of computational resources. Our experiments on "ControlNet Relevance Score", in which we trained a controlled generative model by copying all blocks and removing control blocks from different layers during inference, revealed that different layers in DiT exhibit varying levels of relevance for control information. As shown in Fig. \ref{img:intro_fidandhdd}, this relevance follows a trend of increasing and then decreasing, with higher relevance observed in the front-center layers and lower relevance in the deeper layers of the network, resulting in only a slight performance loss (for a more detailed explanation, please refer to Sec .~\ref {paper: DiT-ControlNet Relevance Prior}). Existing methods neglect this variation and apply uniform settings to all layers that introduce control information, resulting in inefficient allocation of parameters and computational resources, such as redundant parameters or computations in layers with low relevance.

To address the above issues, we propose the Relevance-Guided Efficient Controllable Generation framework (i.e., \textbf{RelaCtrl}) for the diffusion transformer, based on control information relevance analysis. 
Specifically, to achieve efficient utilization of computational resources, we design relevance-guided allocation and steering strategies. Control blocks are placed at locations with high control information relevance, while locations with weak relevance are left without control blocks. 
To further reduce the number of parameters and computational complexity introduced by the copy control block, we design a lightweight Two-Dimensional Shuffle Mixer (\textbf{TDSM}) to replace the self-attention and FFN layers in the copy block.
Self-attention plays the role of token mixer, and FFN plays the role of channel mixer. Therefore, to efficiently perform the same functions as the token mixer and channel mixer, TDSM first randomly selects a varying number of feature channel groups, then randomly divides the token groups within each channel group, and finally computes the attention within each token-channel group along the token dimension.
Theoretical analysis demonstrates that TDSM can overcome the limitations of local grouping and enable non-local modeling in the channel-token dimensions. In addition, we regulate the number of channel division groups in TDSM based on the correlation. In regions with stronger correlation, we reduce the number of channel groups and expand the feature dimensions involved in attention to enhance its modeling capability.
The results from multiple experiments demonstrate that our approach achieves superior performance with only a 45M parameter increase (7.4\% of PixArt-$\alpha$) and an additional 46.7 GFLOPs (8.6\% of PixArt-$\alpha$).

Our contribution can be summarized as follows:
\begin{itemize}
\setlength{\itemsep}{0.5em}    
\setlength{\parskip}{0.3em}    
\item We investigate in detail the relevance of control information across different layers, finding that the shallower layers are more sensitive to the control signal, while the deeper exhibit weaker relevance to the control effect.
\item Based on the relevance analysis, we propose a relevance-guided controlled generation strategy (RelaCtrl), which efficiently allocates the embedding positions of control blocks and the strength of the TDSM modeling capability. This approach minimizes the number of parameters introduced by the control branch and reduces computational complexity without compromising performance.
\item We propose a Two-Dimensional Shuffle Mixer, which efficiently replaces the self-attention and FFN in the original copy block by calculating attention within randomly divided channel and token groups. Theoretical analysis demonstrates that this design overcomes the limitations of local group modeling, ensuring efficient token mixing and channel mixing.
\item The experimental results across different conditional guidance tasks show that RelaCtrl consistently achieves superior performance while maintaining efficiency, validating the generalization of both the relevance-guided strategy and the proposed TDSM.
\end{itemize}

\section{Related Works}
\subsection{Diffusion-Based Models}
In recent years, diffusion-based methods have garnered significant success in the field of generation \cite{he2024idanimator, feng2024fancyvideo, wang2025wisa, zhang2024artbank, bi2024using}, particularly in text-to-image (T2I) generation \cite{guo2023animatediff,shuai2024survey, chen2025ctr, wang2025generate, li2023relation}. These methods utilize text embeddings derived from pre-trained language encoders, such as CLIP \cite{radford2021clip}, BERT \cite{devlin2018bert}, and T5 \cite{raffel2020t5}, to generate images with high fidelity and diversity through an iterative denoising process. Recently, the introduction of the latent diffusion model\cite{rombach2022ldm} has marked a significant advancement in this field, enhancing the quality and efficiency of generated content. In pursuit of greater scalability and enhanced generation quality, models like DiT \cite{peebles2023dit,chen2023PixArta}integrate large-scale Transformer architectures into the diffusion framework, pushing the boundaries of generative performance. Building on this foundation, Flux \cite{blackforestlabs2024flux} synthesizes flow-matching \cite{lipman2022flow} and Transformer-based architecture to achieve state-of-the-art performance.

\subsection{Controllable Generation with Diffusion Models}
Controllable generation has emerged as a prominent area of research in diffusion models \cite{zhang2023controlnet, qin2023unicontrol, zhao2024unicontrolnet, chen2024PixArtc, peng2024controlnext, che2024gamegen}. Currently, there are two main approaches to incorporating controllable conditions into image generation: (1) training a large diffusion model from scratch to enable control under multiple conditions, and (2) fine-tuning a lightweight structure while keeping the original pre-trained model frozen. However, the first approach demands significant computational resources, which limits its accessibility for broader dissemination and personal use. In contrast, recent studies have explored the addition of supplementary network structures to pre-trained diffusion models, allowing for control over the generated outputs without the need to retrain the entire model.
ControlNet~\cite{zhang2023controlnet} enables image generation that aligns with control information by reproducing specific layers within the network and connecting them to the original layers using zero convolution. Building on this foundation, Uni-Control~\cite{qin2023unicontrol} introduces a Mixture-of-Experts (MoE) framework, which unifies control across multiple spatial conditions. ControlNet-XS~\cite{zavadski2025controlnetxs} further improves the interaction bandwidth and frequency between the control and main branches within the ControlNet architecture, drawing inspiration from principles of feedback control systems. Nonetheless, these approaches are primarily based on the U-Net structure and may not yield the desired results when directly applied to Diffusion Transformers (DiT) without modification~\cite{chen2024PixArtc}. PixArt-$\delta$~\cite{chen2024PixArtc} proposed a design methodology specifically tailored for DiT, but directly copying the first half of the network results in a 50\% increase in both parameter count and computational complexity, resulting in high computational cost and being inconvenient for community research and practical deployment.

\section{Methods}
\subsection{DiT-ControlNet Relevance Prior}
\label{paper: DiT-ControlNet Relevance Prior}
In prior studies, the majority of studies on ControlNet have centered on the U-Net architecture. However, the DiT framework~\cite{chen2023PixArta}, constructed by stacking a sequence of transformer blocks without an explicit encoder-decoder structure, poses significant challenges for direct adaptation to achieve effective controllability~\cite{chen2024PixArtc}. To tackle this issue,  PixArt-${\delta}$ introduced a method that duplicates the first 13 Transformer blocks from the DiT model, integrates the outputs of these copied blocks with their corresponding frozen blocks, and forwards the combined results to subsequent frozen modules for further processing. While this approach has demonstrated qualitatively favorable results, it brings notable drawbacks. Simply duplicating the first half of the frozen modules results in a considerable increase in model parameters and computational overhead, leading to prohibitively high training and inference costs, particularly for high-resolution image generation. Moreover, our observations indicate that different replicated layers in DiT-ControlNet contribute unequally to the overall generation quality and control fidelity. Blindly copying the initial Transformer blocks can introduce unnecessary computational redundancy without proportional performance gains.

To systematically evaluate the relevance of individual layers within DiT-ControlNet to generation and controlled quality, we trained a fully controlled PixArt-${\alpha}$ network containing 27 replicated modules. We systematically skip each control block layer during inference and assess its impact on the final generation. We employed the Fréchet Inception Distance (FID) to measure image generation quality and the Hausdorff Distance (HDD) to evaluate control accuracy for quantitative assessment. These metrics enabled us to analyze the impact of skipping individual blocks from the control branch on overall performance, providing relevant scores for each control block. Finally, we get the ControlNet Relevance Score $CRS$ based on the combination of these two metrics:
\begin{equation}\label{equ:cis}
{{CRS}_{i}}=\frac{1}{2}\left( \frac{{{F}_{i}}-{{F}_{\min }}}{{{F}_{\max }}-{{F}_{\min }}}+\frac{{{H}_{i}}-{{H}_{\min }}}{{{H}_{\max }}-{{H}_{\min }}} \right)
\end{equation}
Where  $F$ and $H$ represent the rank of initial FID and HDD indicators, respectively, $i$ shows the index of the control branch block that is removed, and $\min $ and $\max $ denote the minimum and maximum values within the corresponding rank sequence. If ${{F}_{i}}$ or ${{H}_{i}}$ is higher, it indicates that removing the control block with index $i$ significantly affects the final performance, implying that this module is critically important. Using this approach, we performed single-layer deletions across all replicated blocks in the ControlNet model and derived the regularization metrics and qualitative observations presented in Fig.~\ref{img:intro_fidandhdd}. According to the specified formula~\ref{equ:cis}, the relevance distribution of the ControlNet blocks can be obtained. 

Our findings can be summarized as follows. The most critical modules of DiT-ControlNet are concentrated in the early-middle layers (e.g., blocks 5, 6, and 7). In contrast, removing the last few modules results in only a minimal decline in performance. Overall, the ControlNet Relevance Score exhibits a trend of initially increasing and then decreasing, which contrasts with observations from prior studies of large language models \cite{gromov2024unreasonable, zhong2024blockpruner, men2024shortgpt} or the main branches of the original DiT architecture \cite{lee2024ditPruner}. This indicates that simply increasing or decreasing the number of replicated front transformer blocks in DiT-ControlNet does not offer an effective trade-off between performance and computational cost. Such an approach risks removing control modules essential for maintaining performance. Consequently, we propose dynamically guiding the placement and design of control modules within the network by ranking each layer in the DiT model's control branch according to its relevance. This strategy ensures a more targeted and efficient utilization of network resources.
\begin{figure*}[t]
    \centering
    \includegraphics[width=0.95\textwidth]{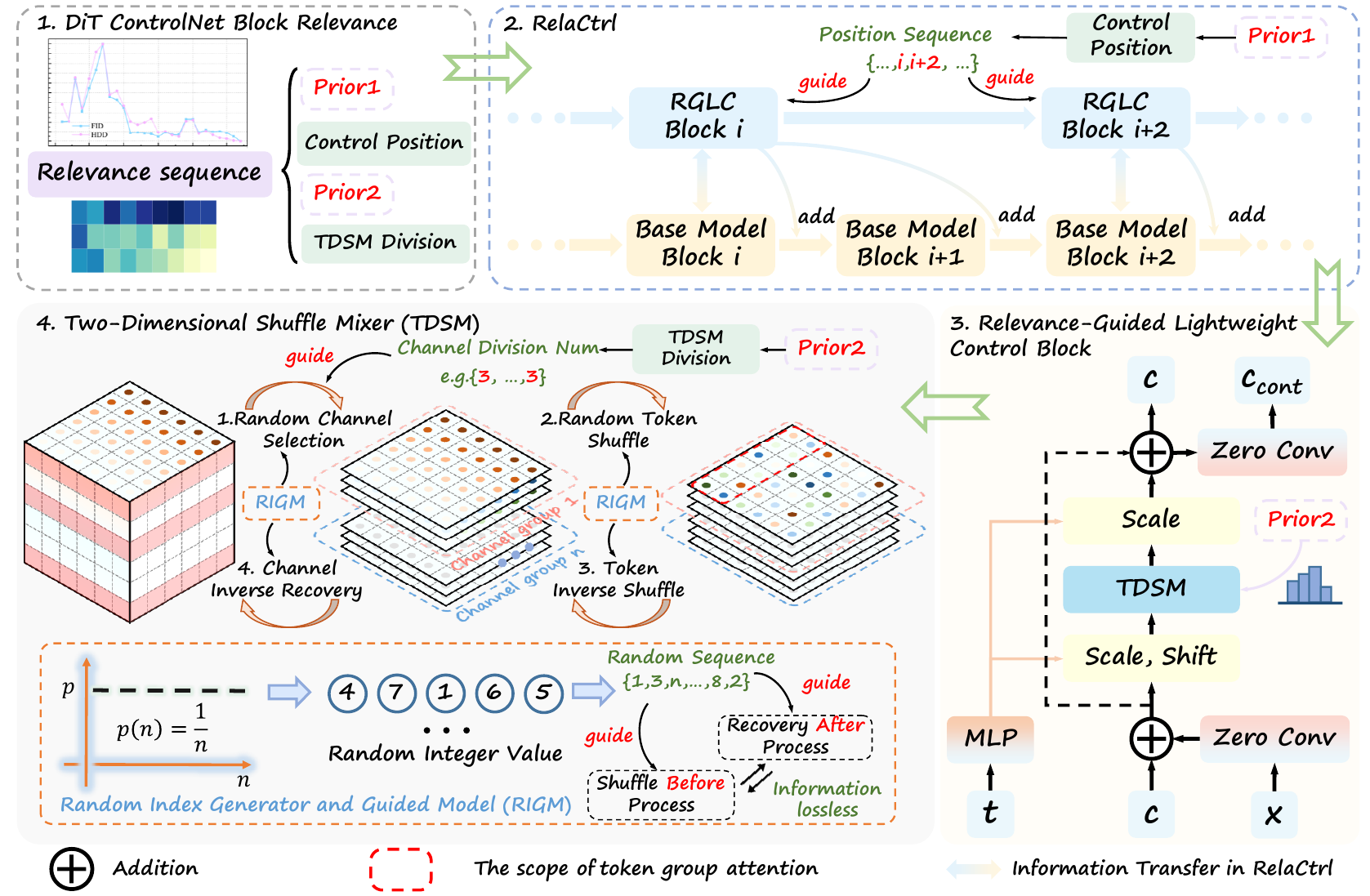}
    \caption{The overall architecture of RelaCtrl. Control block locations are prioritized based on the ControlNet Relevance Score, ranked from highest to lowest. The direct duplication of the main branch in the original ControlNet is replaced with the carefully designed Reference-Guided Lightweight control block. Additionally, the Two-Dimensional Shuffle Mixer effectively reduces model parameters and computational overhead while preserving performance.}
    \label{img:method_arch}
\end{figure*}

\subsection{Overall Architecture}
Fig.~\ref{img:method_arch} depicts the overall pipeline of our proposed method. Based on the ranking of ControlNet Relevance Score derived in Sec .~\ref{paper: DiT-ControlNet Relevance Prior} and further validated through ablation studies, we identified and selected the 11 most critical control positions—ranked by relevance from high to low—for integrating the control modules.
With this approach, we achieve control performance comparable to PixArt-${\delta}$, which utilizes 13 copied modules, while reducing the parameter count by approximately 15\%. Although this method effectively decreases the model size and computational overhead, significant redundancy remains in the internal design of the control modules. The transformative power of the Transformer architecture, as emphasized by MetaFormer~\cite{yu2022metaformer}, lies in its holistic design, wherein the attention mechanism serves as a token mixer by enabling token-level information mixing, while the remaining components, such as the FFN layers, function as channel mixers to facilitate the integration of channel-wise information. To address the substantial redundancy in the FFN layers within the channel mixer~\cite{pires2023sharedFFN}, we introduce a lightweight module, the Relevance-Guided Lightweight Control Block (RGLC), which unifies token mixing and channel mixing into a single operation. Specifically, we replace the attention and FFN layers in the original PixArt Transformer block with a novel Two-Dimensional Shuffle Mixer (TDSM) design, streamlining the architecture for enhanced efficiency. This method facilitates information interaction and modeling across both token and channel dimensions, significantly reducing the replicated blocks' parameter and computational demands.

\subsection{Relevance-Guided Lightweight Control Block}
The third part of Fig.~\ref{img:method_arch} illustrates the detailed structure of the RGLC Block. The module takes three inputs: the control condition input $c$, the diffusion timesteps embedding $t$ used to calculate the weights for feature normalization, and the input $x$ from the corresponding frozen block. To enhance information interaction between the control branch and the frozen backbone network, $x$ is passed through a zero convolution layer and added to the conditional input $c$, producing ${{c}_{in}}$. The resulting ${{c}_{in}}$ is then processed by the Two-Dimensional Shuffle Mixer (TDSM). Following the processing, the output is passed through a zero convolution layer, resulting in ${{c}_{cond}}$, which is added to the main branch to provide control guidance. This process can be formally expressed as follows:
\begin{equation}\label{}
{{c}_{cond}}=\text{ZC}(\text{TDSM}({{c}_{in}})+{{c}_{in}})
\end{equation}
Where $\text{ZC}$ refers to the operation of zero convolution. To address the additional computational overhead introduced by the self-attention mechanism, TDSM employs locally grouped self-attention with shuffle characteristics. This design significantly reduces the computational complexity of the network while preserving non-local information interactions within the groups.
\begin{table*}[!htb]
    \centering
    \small
\begin{tabular}{cc|cccc|cccc}
\toprule[1.1pt]
\multirow{3}{*}{Model} & \multirow{3}{*}{Method}   & Controllability & \multicolumn{2}{c}{Quality}   
& TC & Controllability & \multicolumn{2}{c}{Quality}   
& TC
\\ 
& & \multicolumn{4}{c}{\gb{Canny}} & \multicolumn{4}{c}{\gb{Hed}} \\
& & HDD$\downarrow$ & FID$\downarrow$  & C-Ae$\uparrow$   & C-SC$\uparrow$ & HDD$\downarrow$ & FID$\downarrow$  & C-Ae$\uparrow$   & C-SC$\uparrow$ 
\\ \midrule[1.2pt]
  \multirow{2}{*}{SD1.5} 
& Uni-ControlNet & \underline{95.40}  & 33.81  & 5.207   & 0.259 & \underline{98.78}  & 59.72  & 5.086 & 0.252     \\ 
 & Uni-Control & 97.90 & 91.29  & 4.965 & 0.249 & 100.52 & 91.94  & 4.819 & 0.251   \\ 
 \multirow{2}{*}{SDXL} 
& ControlNet-XS & 101.34  & 21.57  & 5.134 & \textbf{0.286}  & - & - & - & -     \\ 
 & ControlNext & 117.59 & 49.32  & 4.816 & 0.275  & - & - & - & -   \\ 
 \multirow{2}{*}{PixArt-$\alpha$} 
& PixArt-$\delta$ & 96.26 & \underline{21.38}  & \underline{5.508} & 0.279  & 98.91 &\underline{29.22}  & \underline{5.243} & \underline{0.275}    \\ 
 & RelaCtrl & \textbf{94.04}  & \textbf{20.34}  & \textbf{5.584} & \underline{0.282} & \textbf{96.11}  & \textbf{27.73}  & \textbf{5.451} & \textbf{0.276}      \\ 
\bottomrule[1.2pt]
\multirow{2}{*}{Model} & \multirow{2}{*}{Method} & \multicolumn{4}{c}{\gb{Depth}}  & \multicolumn{4}{c}{\gb{Segmentation}} \\
 &    & MSE-d$\downarrow$ & FID$\downarrow$  & C-Ae$\uparrow$   & C-Sc$\uparrow$ & mIoU$\uparrow$ & FID$\downarrow$  & C-Ae$\uparrow$   & C-Sc$\uparrow$   \\ \midrule[1.2pt]
  \multirow{2}{*}{SD1.5} 
& Uni-ControlNet & 102.75  & 43.17  & 5.230 & 0.250 & 0.316  & 40.83  & 5.270 & 0.255      \\ 
 & Uni-Control & 102.46 & 91.94  & 5.327 & 0.249 & \underline{0.382} & 40.74  & 5.462 & 0.258   \\ 
 \multirow{2}{*}{SDXL} 
& ControlNet-XS & \underline{99.20}  & \underline{34.38}  & 5.235 & 0.281  & - & - & - & -     \\ 
 & ControlNext & 101.63 & 73.26  & 4.919 & 0.253  & - & - & - & -   \\ 
 \multirow{2}{*}{PixArt-$\alpha$} 
& PixArt-$\delta$ & 99.69 & 35.21  & \underline{5.723} & \underline{0.283}  & 0.379 & \underline{35.50}  & \underline{5.668} & \underline{0.282}    \\ 
 & RelaCtrl & \textbf{99.11}  & \textbf{33.93}  & \textbf{5.887} & \textbf{0.285} & \textbf{0.405}  & \textbf{33.76}  & \textbf{5.702} & \textbf{0.287}      \\ 
 \bottomrule[1.2pt]
\end{tabular}
\caption{Quantitative comparisons of different methods on the COCO validation set~\cite{lin2014microsoftcoco}. The best results are highlighted in bold, while the second-best results are underlined. TC: Text Consistency, C-SC: CLIP-Score, C-Ae: CLIP-AE.}
\label{table_metric_main}
\end{table*}
\subsection{Two-Dimensional Shuffle Mixer}
\label{sec:tdsm-mixer}
From the perspective of MetaFormer~\cite{yu2022metaformer}, the effectiveness of Transformers can be attributed to two key components: the token mixer, implemented via the self-attention mechanism, and the channel mixer, realized through the feed-forward network (FFN) layer. However, studies have revealed that the FFN is often highly redundant despite consuming a significant proportion of the model parameters~\cite{pires2023sharedFFN}. To alleviate the computational burden of the control branch, we propose grouping tokens for computation while employing specific strategies to enhance interaction and modeling capacity across token groups ~\cite{huang2021shuffle,cao2024shufflemamba}. Departing from traditional shuffle methods that operate exclusively on the token dimension, we extend the token mixer in the Transformer architecture to model non-local interactions within local windows by introducing a novel approach that jointly operates on both the token and channel dimensions. This dual-dimension strategy enables more efficient and effective modeling. Specifically, we perform random channel selection, followed by random shuffling of the input sequence across the 3D dimension space. Afterward, local self-attention calculations are applied. Although the subsequent attention mechanism is confined to a fixed group, the tokens involved may originate outside this group. This operation effectively disrupts the inherent relationships between tokens and introduces the information flow between channels to some extent, thereby breaking the interaction constraints typically imposed by local attention. To provide theoretical validation, we first present the following definition:
\begin{definition}
\label{def:Local 3D Partition}
(Local Partition). For the local group where self-attention calculations are performed, its 3D size $s\times s \times d$ satisfies $s\ll H$, $s\ll W$, and $d\ll D$, where $H\times W\times D$ denotes the dimensions of the module input after the token positions have been arranged.
\end{definition}
\begin{definition}
\label{def: Random Selection and Shuffle Function}
(Random Selection and Shuffle Function). $S:{{c}_{in}}\to c_{rs}^{i},i\in [1,n]$ denote the random selection and shuffle function. This function randomly divides the input ${{c}_{in}}\in {{\mathbb{R}}^{H\times W\times D}}$ into $n$ parts along the channel and then scrambles the elements within each part in the 3D space. As a result, $c_{rs}^{i}\in {{\mathbb{R}}^{H\times W\times {{d}_{i}}}},i\in [1,n]$, where $\sum\limits_{i=1}^{n}{{{d}_{i}}}=D$. 

Regardless of the initial distances between tokens, the random selection and shuffle operations may place any two tokens within the same window, making it possible to model non-local relationships at both the token and channel levels within local windows.
\end{definition}
\begin{definition}
\label{def: Interactive Set and Interactive distance}
(Interactive Set and Interactive Distance). Consider $c_{rs}^{i}\in {{\mathbb{R}}^{H\times W\times {{d}_{i}}}}$, where we define the set of token pairs within the same local group after the random shuffle operation as ${{I}_{S}}=\{({{t}_{j}},{{t}_{k}})\}$. This set indicates that, after the random shuffle $S$, the token indices ${{t}_{j}}$ and ${{t}_{k}}$ are placed within the same local group. At this point, the interactive distance between the two tokens can be defined as ${{d}_{S}}$:
\end{definition}
\begin{equation}\label{}
    {{d}_{S}}({{t}_{j}},{{t}_{k}}) =
    \begin{cases}
        {{\left\| {{t}_{j}}-{{t}_{k}} \right\|}_{2}}, & \text{if} \, ({{t}_{j}},{{t}_{k}}) \in {{I}_{S}} \\
        \infty, & \text{if} \, ({{t}_{j}},{{t}_{k}}) \notin {{I}_{S}}
    \end{cases}
\end{equation}
Here, $\infty$ indicates that the interaction between the two tokens cannot be captured within the current local group, which does not affect the derivation process of distance modeling. Therefore, the expected interactive distance $d({{t}_{j}})$ of token ${{t}_{j}}$ can be defined as follows:
\begin{equation}\label{}
d({{t}_{j}})=\underset{\left. {{t}_{k}} \right|({{t}_{j}},{{t}_{k}})\in {{I}_{S}}}{\mathop{\mathbb{E}}}\,\left[ {{d}_{S}}({{t}_{j}},{{t}_{k}}) \right]
\end{equation}
It can be proven that the lower bound of $d({{t}_{j}})$ is $\Omega (\frac{\sqrt{2}}{4}(H+W{{d}_{i}}))$ . 
\begin{theorem}
\label{thm: lower bound}
The lower bound of $d({{t}_{j}})$ is:
\end{theorem}
\begin{equation}
\begin{split}
  d({{t}_{j}}) &\ge \frac{\sqrt{2}}{4(HW{{d}_{i}}-1)} \Big[ H{{t}_{wj}}({{t}_{wj}}+1) \\
  & \quad + W{{d}_{i}}{{t}_{hj}}({{t}_{hj}}+1) \\
  & \quad + H(W{{d}_{i}}-{{t}_{wj}})(W{{d}_{i}}-{{t}_{wj}}-1) \\
  & \quad + W{{d}_{i}}(H-{{t}_{hj}})(H-{{t}_{hj}}-1) \Big]
\end{split}
\end{equation}
\begin{corollary}
\label{cor2.5}
Let $\bar{d}({{t}_{j}})$ denote the average interactive distance in $c_{rs}^{i}$, and it can be proved that $\bar{d}({{t}_{j}})\approx d({{t}_{j}})$.
\end{corollary}
According to Corollary~\ref{cor2.5}, the average distance that can be captured by grouped attention in TDSM is $\Omega (\frac{\sqrt{2}}{4}(H+W{{d}_{i}}))$, enabling the modeling of non-local interactions. In summary, the random selection in the channel and the shuffle operation we introduced disrupt the positional order of tokens. However, since the token order is not explicitly modeled within the self-attention mechanism for visual tasks, this operation does not compromise the effectiveness of the process. Nonetheless, the arrangement of input tokens may affect the semantic information embedded in the latent code during recovery. To resolve this problem, we propose an inverse recovery operation applied to both the token and channel dimensions following the self-attention computation. This overall method with shuffle and recovery is termed the Two-Dimensional Shuffle Mixer (TDSM), leveraging the capability of these reversible transformation pairs to ensure information preservation during self-attention calculations, thereby enabling efficient non-local information interaction across both the channel and token dimensions.
\section{Experiment}
\label{paper: Experiment}
\subsection{Experiment Setup}
\textbf{Evaluation Metrics.}To comprehensively assess the quality of the generated images, we employed multiple evaluation metrics. The Fréchet Inception Distance (FID) \cite{heusel2017fid} and CLIP-Aesthetics score (C-Ae)~\cite{schuhmann2022clipae}measure the visual fidelity of the generated images, while the CLIP Score (C-SC)~\cite{hessel2021clipscore} evaluates text consistency. The control fidelity is explicitly assessed through the distance between the generated and target images. For images guided by HED and Canny conditions, we utilized the Hausdorff Distance (HDD) \cite{huttenlocher1993comparing} for evaluation. For depth conditions, we used MSE-depth, and for segmentation map control, we employed the mean Intersection over Union (mIoU).

\textbf{Baseline.} We conducted a comparative analysis of RelaCtrl against state-of-the-art (SOTA) control techniques, including Uni-ControlNet \cite{zhao2024unicontrolnet}, UniControl \cite{qin2023unicontrol}, ControlNet-XS \cite{zavadski2025controlnetxs}, ControlNext \cite{peng2024controlnext}, and PixArt-$\delta$ \cite{chen2024PixArtc}. For the first four methods, we utilized their officially released pre-trained weights. It is worth noting that ControlNet-XS and ControlNext only provided weights for Canny and Depth conditions, with the Canny weights for ControlNext being trained on an animation dataset. To further highlight the performance of our proposed method, we trained the control branch for both PixArt-$\delta$ and our method entirely from scratch under identical experimental settings, enabling a rigorous quantitative and qualitative comparison.
\begin{figure}[t]
\centering
\includegraphics[width=\linewidth]{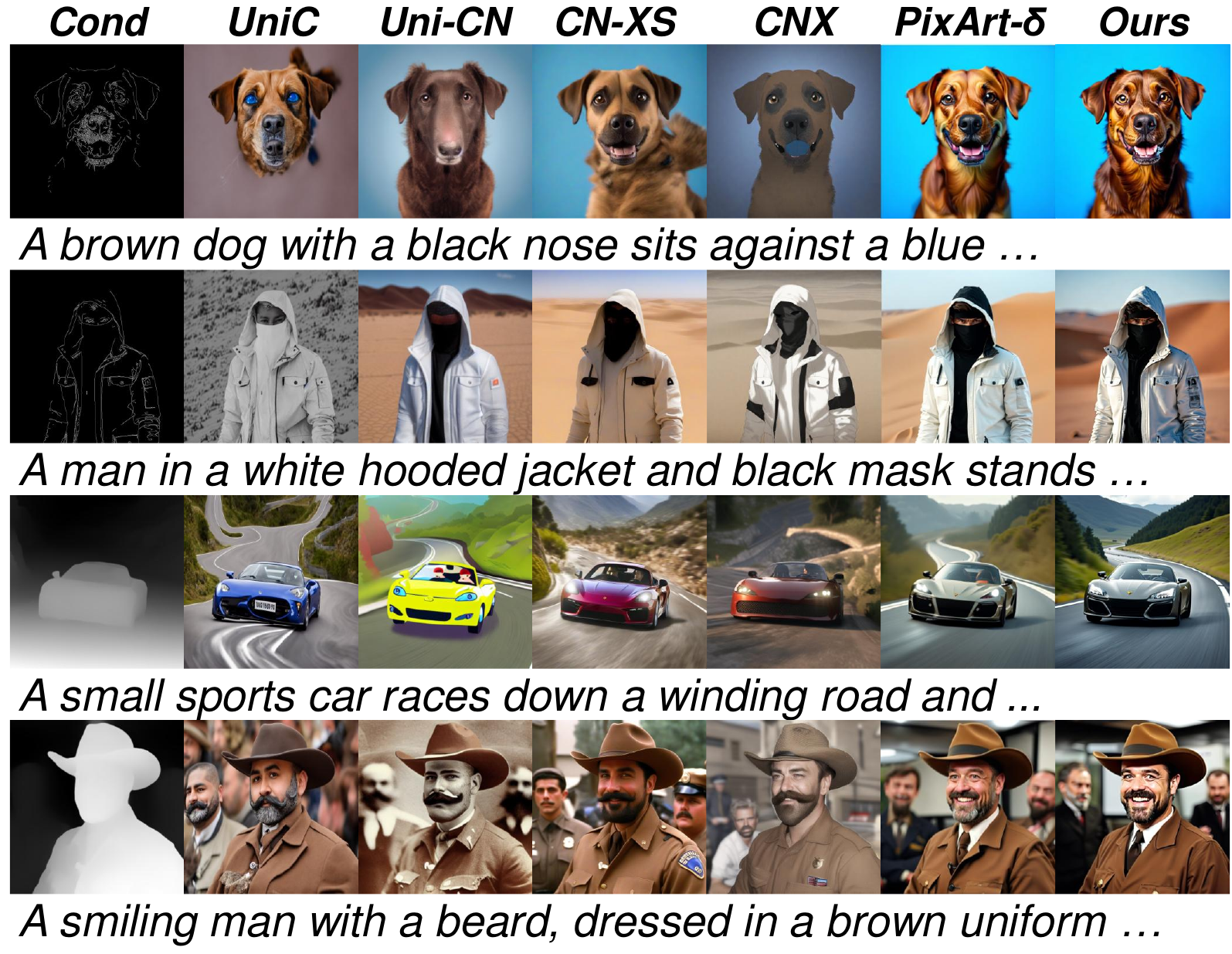}
\caption{\label{img:exp_compare}Qualitative comparison of different methods. From left to right, each row shows the conditioning input, UniControl, Uni-ControlNet, ControlNet-XS, ControlNext, PixArt-$\delta$, and our RelaCtrl.
}
\end{figure}
\subsection{Compared with SOTA Methods}
As shown in Table \ref{table_metric_main}, we comprehensively evaluated the proposed method against existing controllable generation techniques across four conditional control tasks. Our method consistently outperforms alternatives in control accuracy, as demonstrated by the superior performance on control indicators for various conditions, highlighting its precision in generating controlled images. Furthermore, our approach achieves consistently better results in terms of FID and CLIP aesthetic scores, reflecting enhanced image quality. In text similarity evaluations, the CLIP Score confirms that our method achieves superior text-image consistency across diverse tasks, demonstrating improved semantic alignment while maintaining high control accuracy and visual fidelity. 

\begin{figure}[t]
\centering
\includegraphics[width=\linewidth]{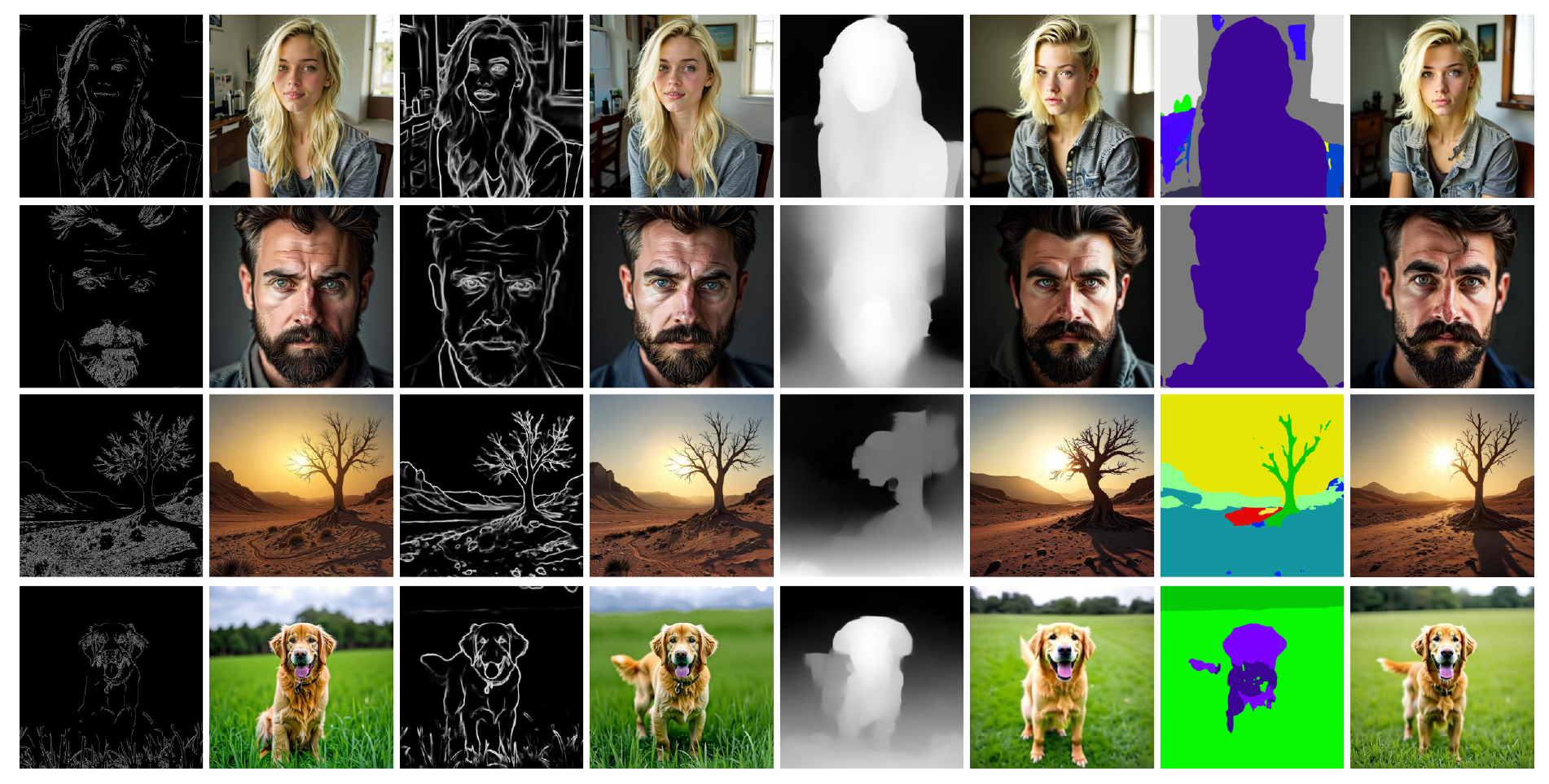}
\caption{\label{img:exp_visual}Generation effects of RelaCtrl under varying control conditions.
}

\end{figure}

Fig.~\ref{img:exp_compare} presents visual comparisons of RelaCtrl and other methods under Canny and Depth conditions. Our method significantly reduces computational complexity and parameter usage while maintaining generation quality comparable to PixArt-ControlNet, outperforming other approaches. Additional controllable generation results of RelaCtrl are shown in Fig.~\ref{img:exp_visual}, further illustrating its effectiveness in extracting and injecting information from conditional images during the generation process. This enables the production of results that align seamlessly with conditional controls.
\subsection{Algorithm Efficiency Analysis}
We conducted a comprehensive analysis of RelaCtrl and ControlNet with 13 replicated modules (PixArt-$\delta$) under consistent experimental settings. Specifically, the input data was in bfloat16 format with a resolution of 512, the test batch size was set to 1, and all experiments were performed on an NVIDIA A100 GPU. Table~\ref{tab: effective} presents a comparative evaluation of the two methods, with each indicator expressed as a percentage relative to the original base model. Notably, memory consumption excludes usage by the CLIP and T5 encoders. Compared to the original PixArt-$\alpha$, RelaCtrl incurred a modest 7.38\% increase in parameters and an 8.61\% increase in computational complexity. These increments are significantly lower than those of the ControlNet method, which demonstrates nearly a 50\% increase in both parameters and complexity. Additionally, RelaCtrl showed some advantages in inference time, although it is worth noting that inference time is predominantly influenced by the speed of the main network rather than the control branch, leading to similar small increases in this metric for both methods. Overall, RelaCtrl outperforms the original ControlNet method applied to the DiT model by achieving comparable or superior metrics and visual results while significantly reducing computational resource consumption. This demonstrates the effectiveness and efficiency of the proposed RelaCtrl framework.
\begin{table}[t]
    \centering
    \small
\begin{tabular}{cccccc}
\hline
 Method   & Parameters & Complexity & Inference  \\ \hline
    PixArt-$\alpha$
  & 611.15 & 542.56  & 3.81    \\ 
   \multirow{2}{*}{w/ ControlNet} 
  & +294.34 & +270.57  & +0.51    \\ 
  &(+48.16\%)  & (+49.87\%)  & (+13.39\%) \\
 \multirow{2}{*}{w/ RelaCtrl} 
  & +\textbf{45.15}  & +\textbf{46.71}  & +\textbf{0.24}    \\ 
  &(+\textbf{7.38\%})  & (+\textbf{8.61\%})  & (+\textbf{6.30\%}) \\
\hline
\end{tabular}
\caption{Evaluation of the proposed method's effectiveness, with the following units for the three metrics: Parameters (M), Complexity (GFLOPs), Inference Time (s).}
    \label{tab: effective}
\end{table}
\subsection{Ablation Study}
We conducted quantitative experiments to evaluate the efficacy of control block placement based on ControlNet Relevance. Utilizing relevance scores, we ranked the control block positions from highest to lowest and placed copy blocks at the top-ranked positions, evaluating configurations with 13, 12, 11, and 10 blocks. The results are summarized in Table~\ref{tab:abl1}. As anticipated, reducing the number of control blocks resulted in a gradual decline in FID and HDD metrics. Notably, under the guidance of relevance scores, using only 11 control blocks achieved performance comparable to the original ControlNet with 13 replicated blocks. When further reduced to 10 blocks, the quality of the generated results slightly declined relative to the 13-block setup. Consequently, we selected the top 11 positions ranked by relevance scores for control placement.

To further validate the significance of the RGLC block and the incorporation of Prior 2 based on revelation, we conducted additional quantitative experiments, which are shown in Table~\ref{tab:abl2}. Employing the ControlNet with 13 copy blocks as the baseline and the RelaCtrl network as the foundation, we evaluated the impact of removing the RGLC block and relevance prior 2, respectively. The results demonstrate that the absence of either component leads to a decline in image quality and control performance. Replacing the RGLC block with the original copy block not only significantly increases the parameter scale but also results in performance degradation, highlighting the efficacy of the RGLC block. In the experiment (w/o Prior 2), we applied a uniform TDSM channel division across all RGLC blocks and eliminated our correlation-specific settings. The poorer HDD and FID metrics observed indicate that allocating more parameters and computational resources to control positions with higher correlation is crucial for achieving high-quality controllable generation. These results underscore the importance of both the RGLC block and relevance prior 2 in enhancing generation quality and control precision.
\begin{table}[t]
    \centering
    \small
\begin{tabular}{cccccccc}
\hline
 Setting   & HDD$\downarrow$ & FID$\downarrow$  & Para Ratio
\\ \hline
 ControlNet-top13 & 96.26  & 21.38   & 100\% \\ 
 Relevance-top13  & 94.57  & 20.31   & 100\%  \\ 
 Relevance-top12  & 95.88  & 20.79   & 92.5\%   \\  
 \rowcolor{gray!20}
 Relevance-top11  & 95.57  & 21.28   & 84.6\% \\ 
 Relevance-top10  & 96.36  & 22.24   & 76.9\% \\ 
\hline
\end{tabular}
\caption{The impact of control block placement guided by DiT-ControlNet Relevance. ControlNet-top13, which directly replicates the first 13 blocks of the main branch, serves as the baseline for parameter volume comparison.}
    \label{tab:abl1}
\end{table}
\begin{table}[!t]
    \centering
    \small
\begin{tabular}{cccccccc}
\hline
 Setting   & HDD$\downarrow$ & FID$\downarrow$ & Para Ratio
\\ \hline
 \rowcolor{gray!20}
 RelaCtrl      & 94.04  & 20.34    & 15.3\%  \\ 
 w/o RGLC      & 95.57  & 21.28    & 84.6\% \\  
 w/o Prior2    & 97.30  & 22.47    & 17.1\%  \\ 
 Baseline      & 96.26  & 21.38    & 100\%  \\ 
\hline
\end{tabular}
\caption{The impact of the RGLC block and the number of TDSM partitions within it on generation performance. The PixArt-$\delta$ with 13 copied blocks serves as the baseline for parameter comparison.}
    \label{tab:abl2}
\end{table}
\section{Conclusion}
In this paper, we explore the control information relevance across different layers in the diffusion transformer by the "ControlNet Relevance Score" experiments. We discover that layers with strong relevance to control information are located in the shallow to middle layers, while the deeper layers exhibit weaker relevance. Then, we propose a relevance-guided strategy (RelaCtrl) for allocating control block insertion positions and adjusting the block's modeling capabilities, enhancing the efficiency of control information integration. Additionally, we design TDSM, which efficiently replaces the original self-attention and FFN through a randomized channel-token grouping attention mechanism.
Experimental results show that RelaCtrl achieves superior performance across two T2I models and four conditional guidance generation tasks, while also exhibiting significant efficiency advantages. We hope RelaCtrl provides valuable insights and references for controlled generation research based on diffusion transformers.

\section{Acknowledgments}
This work was supported by the Hefei Municipal Natural Science Foundation, HZR2417.
\bibliography{aaai2026}

@inproceedings{chen2025ctr,
  title={CTR-driven advertising image generation with multimodal large language models},
  author={Chen, Xingye and Feng, Wei and Du, Zhenbang and Wang, Weizhen and Chen, Yanyin and Wang, Haohan and Liu, Linkai and Li, Yaoyu and Zhao, Jinyuan and Li, Yu and others},
  booktitle={Proceedings of the ACM on Web Conference 2025},
  pages={2262--2275},
  year={2025}
}

@inproceedings{wang2025generate,
  title={Generate e-commerce product background by integrating category commonality and personalized style},
  author={Wang, Haohan and Feng, Wei and Li, Yaoyu and Zhang, Zheng and Lv, Jingjing and Shen, Junjie and Lin, Zhangang and Shao, Jingping},
  booktitle={ICASSP 2025-2025 IEEE International Conference on Acoustics, Speech and Signal Processing (ICASSP)},
  pages={1--5},
  year={2025},
  organization={IEEE}
}

@inproceedings{li2023relation,
  title={Relation-aware diffusion model for controllable poster layout generation},
  author={Li, Fengheng and Liu, An and Feng, Wei and Zhu, Honghe and Li, Yaoyu and Zhang, Zheng and Lv, Jingjing and Zhu, Xin and Shen, Junjie and Lin, Zhangang and others},
  booktitle={Proceedings of the 32nd ACM International Conference on Information and Knowledge Management},
  pages={1249--1258},
  year={2023}
}

@inproceedings{yu2022metaformer,
  title={Metaformer is actually what you need for vision},
  author={Yu, Weihao and Luo, Mi and Zhou, Pan and Si, Chenyang and Zhou, Yichen and Wang, Xinchao and Feng, Jiashi and Yan, Shuicheng},
  booktitle={Proceedings of the IEEE/CVF conference on computer vision and pattern recognition},
  pages={10819--10829},
  year={2022}
}

@article{pires2023sharedFFN,
  title={One wide feedforward is all you need},
  author={Pires, Telmo Pessoa and Lopes, Ant{\'o}nio V and Assogba, Yannick and Setiawan, Hendra},
  journal={arXiv preprint arXiv:2309.01826},
  year={2023}
}

@inproceedings{lee2024ditPruner,
  title={DiT-Pruner: Pruning Diffusion Transformer Models for Text-to-Image Synthesis Using Human Preference Scores},
  author={Lee, Youngwan and Lee, Yong-Ju and Hwang, Sung Ju},
  booktitle={European Conference on Computer Vision (ECCV) 2024},
  pages={1--9},
  year={2024}
}

@article{gromov2024unreasonable,
  title={The unreasonable ineffectiveness of the deeper layers},
  author={Gromov, Andrey and Tirumala, Kushal and Shapourian, Hassan and Glorioso, Paolo and Roberts, Daniel A},
  journal={arXiv preprint arXiv:2403.17887},
  year={2024}
}

@article{zhong2024blockpruner,
  title={BlockPruner: Fine-grained Pruning for Large Language Models},
  author={Zhong, Longguang and Wan, Fanqi and Chen, Ruijun and Quan, Xiaojun and Li, Liangzhi},
  journal={arXiv preprint arXiv:2406.10594},
  year={2024}
}

@article{wang2024qihoo,
  title={Qihoo-T2X: An Efficient Proxy-Tokenized Diffusion Transformer for Text-to-Any-Task},
  author={Wang, Jing and Ma, Ao and Feng, Jiasong and Leng, Dawei and Yin, Yuhui and Liang, Xiaodan},
  journal={arXiv preprint arXiv:2409.04005},
  year={2024}
}

@article{he2024idanimator,
  title={Id-animator: Zero-shot identity-preserving human video generation},
  author={He, Xuanhua and Liu, Quande and Qian, Shengju and Wang, Xin and Hu, Tao and Cao, Ke and Yan, Keyu and Zhang, Jie},
  journal={arXiv preprint arXiv:2404.15275},
  year={2024}
}

@article{cao2024shufflemamba,
  title={Shuffle mamba: State space models with random shuffle for multi-modal image fusion},
  author={Cao, Ke and He, Xuanhua and Hu, Tao and Xie, Chengjun and Zhang, Jie and Zhou, Man and Hong, Danfeng},
  journal={arXiv preprint arXiv:2409.01728},
  year={2024}
}

@inproceedings{zhang2023controlnet,
  title={Adding conditional control to text-to-image diffusion models},
  author={Zhang, Lvmin and Rao, Anyi and Agrawala, Maneesh},
  booktitle={Proceedings of the IEEE/CVF International Conference on Computer Vision},
  pages={3836--3847},
  year={2023}
}

@article{qin2023unicontrol,
  title={Unicontrol: A unified diffusion model for controllable visual generation in the wild},
  author={Qin, Can and Zhang, Shu and Yu, Ning and Feng, Yihao and Yang, Xinyi and Zhou, Yingbo and Wang, Huan and Niebles, Juan Carlos and Xiong, Caiming and Savarese, Silvio and others},
  journal={arXiv preprint arXiv:2305.11147},
  year={2023}
}

@article{zhao2024unicontrolnet,
  title={Uni-controlnet: All-in-one control to text-to-image diffusion models},
  author={Zhao, Shihao and Chen, Dongdong and Chen, Yen-Chun and Bao, Jianmin and Hao, Shaozhe and Yuan, Lu and Wong, Kwan-Yee K},
  journal={Advances in Neural Information Processing Systems},
  volume={36},
  year={2024}
}

@inproceedings{zavadski2025controlnetxs,
  title={ControlNet-XS: Rethinking the Control of Text-to-Image Diffusion Models as Feedback-Control Systems},
  author={Zavadski, Denis and Feiden, Johann-Friedrich and Rother, Carsten},
  booktitle={European Conference on Computer Vision},
  pages={343--362},
  year={2025},
  organization={Springer}
}

@article{peng2024controlnext,
  title={Controlnext: Powerful and efficient control for image and video generation},
  author={Peng, Bohao and Wang, Jian and Zhang, Yuechen and Li, Wenbo and Yang, Ming-Chang and Jia, Jiaya},
  journal={arXiv preprint arXiv:2408.06070},
  year={2024}
}

@article{tan2024ominicontrol,
  title   = {OminiControl: Minimal and Universal Control for Diffusion Transformer},
  author  = {Zhenxiong Tan and Songhua Liu and Xingyi Yang and Qiaochu Xue and Xinchao Wang},
  journal = {arXiv preprint arXiv:2411.15098},
  year    = {2024}
}

@inproceedings{lin2014microsoftcoco,
  title={Microsoft coco: Common objects in context},
  author={Lin, Tsung-Yi and Maire, Michael and Belongie, Serge and Hays, James and Perona, Pietro and Ramanan, Deva and Doll{\'a}r, Piotr and Zitnick, C Lawrence},
  booktitle={Computer Vision--ECCV 2014: 13th European Conference, Zurich, Switzerland, September 6-12, 2014, Proceedings, Part V 13},
  pages={740--755},
  year={2014},
  organization={Springer}
}

@article{heusel2017fid,
  title={Gans trained by a two time-scale update rule converge to a local nash equilibrium},
  author={Heusel, Martin and Ramsauer, Hubert and Unterthiner, Thomas and Nessler, Bernhard and Hochreiter, Sepp},
  journal={Advances in neural information processing systems},
  volume={30},
  year={2017}
}

@article{hessel2021clipscore,
  title={Clipscore: A reference-free evaluation metric for image captioning},
  author={Hessel, Jack and Holtzman, Ari and Forbes, Maxwell and Bras, Ronan Le and Choi, Yejin},
  journal={arXiv preprint arXiv:2104.08718},
  year={2021}
}

@article{schuhmann2022clipae,
  title={Laion-5b: An open large-scale dataset for training next generation image-text models},
  author={Schuhmann, Christoph and Beaumont, Romain and Vencu, Richard and Gordon, Cade and Wightman, Ross and Cherti, Mehdi and Coombes, Theo and Katta, Aarush and Mullis, Clayton and Wortsman, Mitchell and others},
  journal={Advances in Neural Information Processing Systems},
  volume={35},
  pages={25278--25294},
  year={2022}
}

@misc{stabilityai2024sd3,
  author = {{Stability AI}},
  title = {StableDiffusion3},
  howpublished = {\url{https://stability.ai/news/stable-diffusion-3}},
  year = {2024},
  note = {Accessed: 2024-09-03}
}

@article{chen2023pixarta,
  title={Pixart-$\alpha$: Fast training of diffusion transformer for photorealistic text-to-image synthesis},
  author={Chen, Junsong and Yu, Jincheng and Ge, Chongjian and Yao, Lewei and Xie, Enze and Wu, Yue and Wang, Zhongdao and Kwok, James and Luo, Ping and Lu, Huchuan and others},
  journal={arXiv preprint arXiv:2310.00426},
  year={2023}
}

@article{chen2024pixartc,
  title={Pixart-$\delta$: Fast and controllable image generation with latent consistency models},
  author={Chen, Junsong and Wu, Yue and Luo, Simian and Xie, Enze and Paul, Sayak and Luo, Ping and Zhao, Hang and Li, Zhenguo},
  journal={arXiv preprint arXiv:2401.05252},
  year={2024}
}

@misc{blackforestlabs2024flux,
  author = {{BlackForestlabs AI}},
  title = {Flux},
  howpublished = {\url{https://blackforestlabs.ai/#get-flux}},
  year = {2024},
  note = {Accessed: 2024-09-03}
}

@article{yang2024cogvideox,
  title={Cogvideox: Text-to-video diffusion models with an expert transformer},
  author={Yang, Zhuoyi and Teng, Jiayan and Zheng, Wendi and Ding, Ming and Huang, Shiyu and Xu, Jiazheng and Yang, Yuanming and Hong, Wenyi and Zhang, Xiaohan and Feng, Guanyu and others},
  journal={arXiv preprint arXiv:2408.06072},
  year={2024}
}

@inproceedings{radford2021clip,
  title={Learning transferable visual models from natural language supervision},
  author={Radford, Alec and Kim, Jong Wook and Hallacy, Chris and Ramesh, Aditya and Goh, Gabriel and Agarwal, Sandhini and Sastry, Girish and Askell, Amanda and Mishkin, Pamela and Clark, Jack and others},
  booktitle={International conference on machine learning},
  pages={8748--8763},
  year={2021},
  organization={PMLR}
}

@article{devlin2018bert,
  title={Bert: Pre-training of deep bidirectional transformers for language understanding},
  author={Devlin, Jacob},
  journal={arXiv preprint arXiv:1810.04805},
  year={2018}
}

@article{raffel2020t5,
  title={Exploring the limits of transfer learning with a unified text-to-text transformer},
  author={Raffel, Colin and Shazeer, Noam and Roberts, Adam and Lee, Katherine and Narang, Sharan and Matena, Michael and Zhou, Yanqi and Li, Wei and Liu, Peter J},
  journal={Journal of machine learning research},
  volume={21},
  number={140},
  pages={1--67},
  year={2020}
}

@inproceedings{peebles2023dit,
  title={Scalable diffusion models with transformers},
  author={Peebles, William and Xie, Saining},
  booktitle={Proceedings of the IEEE/CVF International Conference on Computer Vision},
  pages={4195--4205},
  year={2023}
}

@article{videoworldsimulators2024,
  title={Video generation models as world simulators},
  author={Tim Brooks and Bill Peebles and Connor Holmes and Will DePue and Yufei Guo and Li Jing and David Schnurr and Joe Taylor and Troy Luhman and Eric Luhman and Clarence Ng and Ricky Wang and Aditya Ramesh},
  year={2024},
  url={https://openai.com/research/video-generation-models-as-world-simulators},
}

@article{li2024hunyuan,
  title={Hunyuan-DiT: A Powerful Multi-Resolution Diffusion Transformer with Fine-Grained Chinese Understanding},
  author={Li, Zhimin and Zhang, Jianwei and Lin, Qin and Xiong, Jiangfeng and Long, Yanxin and Deng, Xinchi and Zhang, Yingfang and Liu, Xingchao and Huang, Minbin and Xiao, Zedong and others},
  journal={arXiv preprint arXiv:2405.08748},
  year={2024}
}

@article{lipman2022flow,
  title={Flow matching for generative modeling},
  author={Lipman, Yaron and Chen, Ricky TQ and Ben-Hamu, Heli and Nickel, Maximilian and Le, Matt},
  journal={arXiv preprint arXiv:2210.02747},
  year={2022}
}

@article{guo2023animatediff,
  title={Animatediff: Animate your personalized text-to-image diffusion models without specific tuning},
  author={Guo, Yuwei and Yang, Ceyuan and Rao, Anyi and Liang, Zhengyang and Wang, Yaohui and Qiao, Yu and Agrawala, Maneesh and Lin, Dahua and Dai, Bo},
  journal={arXiv preprint arXiv:2307.04725},
  year={2023}
}

@inproceedings{zhang2024artbank,
  title={ArtBank: Artistic Style Transfer with Pre-trained Diffusion Model and Implicit Style Prompt Bank},
  author={Zhang, Zhanjie and Zhang, Quanwei and Xing, Wei and Li, Guangyuan and Zhao, Lei and Sun, Jiakai and Lan, Zehua and Luan, Junsheng and Huang, Yiling and Lin, Huaizhong},
  booktitle={Proceedings of the AAAI Conference on Artificial Intelligence},
  volume={38},
  number={7},
  pages={7396--7404},
  year={2024}
}

@article{feng2024fancyvideo,
  title={FancyVideo: Towards Dynamic and Consistent Video Generation via Cross-frame Textual Guidance},
  author={Feng, Jiasong and Ma, Ao and Wang, Jing and Cheng, Bo and Liang, Xiaodan and Leng, Dawei and Yin, Yuhui},
  journal={arXiv preprint arXiv:2408.08189},
  year={2024}
}

@inproceedings{rombach2022ldm,
  title={High-resolution image synthesis with latent diffusion models},
  author={Rombach, Robin and Blattmann, Andreas and Lorenz, Dominik and Esser, Patrick and Ommer, Bj{\"o}rn},
  booktitle={Proceedings of the IEEE/CVF conference on computer vision and pattern recognition},
  pages={10684--10695},
  year={2022}
}

@article{shuai2024survey,
  title={A Survey of Multimodal-Guided Image Editing with Text-to-Image Diffusion Models},
  author={Shuai, Xincheng and Ding, Henghui and Ma, Xingjun and Tu, Rongcheng and Jiang, Yu-Gang and Tao, Dacheng},
  journal={arXiv preprint arXiv:2406.14555},
  year={2024}
}

@article{huttenlocher1993comparing,
  title={Comparing images using the Hausdorff distance},
  author={Huttenlocher, Daniel P and Klanderman, Gregory A. and Rucklidge, William J},
  journal={IEEE Transactions on pattern analysis and machine intelligence},
  volume={15},
  number={9},
  pages={850--863},
  year={1993},
  publisher={IEEE}
}

@article{men2024shortgpt,
  title={Shortgpt: Layers in large language models are more redundant than you expect},
  author={Men, Xin and Xu, Mingyu and Zhang, Qingyu and Wang, Bingning and Lin, Hongyu and Lu, Yaojie and Han, Xianpei and Chen, Weipeng},
  journal={arXiv preprint arXiv:2403.03853},
  year={2024}
}

@article{huang2021shuffle,
  title={Shuffle transformer: Rethinking spatial shuffle for vision transformer},
  author={Huang, Zilong and Ben, Youcheng and Luo, Guozhong and Cheng, Pei and Yu, Gang and Fu, Bin},
  journal={arXiv preprint arXiv:2106.03650},
  year={2021}
}

@article{wang2025wisa,
  title={Wisa: World simulator assistant for physics-aware text-to-video generation},
  author={Wang, Jing and Ma, Ao and Cao, Ke and Zheng, Jun and Zhang, Zhanjie and Feng, Jiasong and Liu, Shanyuan and Ma, Yuhang and Cheng, Bo and Leng, Dawei and others},
  journal={arXiv preprint arXiv:2503.08153},
  year={2025}
}

@inproceedings{ma2025lay2story,
  title={Lay2Story: Extending Diffusion Transformers for Layout-Togglable Story Generation},
  author={Ma, Ao and Feng, Jiasong and Cao, Ke and Wang, Jing and Wang, Yun and Zhang, Quanwei and Zhang, Zhanjie},
  booktitle={Proceedings of the IEEE/CVF International Conference on Computer Vision},
  pages={16102--16111},
  year={2025}
}

@inproceedings{he2025plangen,
  title={Plangen: Towards unified layout planning and image generation in auto-regressive vision language models},
  author={He, Runze and Cheng, Bo and Ma, Yuhang and Jia, Qingxiang and Liu, Shanyuan and Ma, Ao and Wu, Xiaoyu and Wu, Liebucha and Leng, Dawei and Yin, Yuhui},
  booktitle={Proceedings of the IEEE/CVF International Conference on Computer Vision},
  pages={18143--18154},
  year={2025}
}

@inproceedings{wang2025learning,
  title={Learning robust stereo matching in the wild with selective mixture-of-experts},
  author={Wang, Yun and Wang, Longguang and Zhang, Chenghao and Zhang, Yongjian and Zhang, Zhanjie and Ma, Ao and Fan, Chenyou and Lam, Tin Lun and Hu, Junjie},
  booktitle={Proceedings of the IEEE/CVF International Conference on Computer Vision},
  pages={21276--21287},
  year={2025}
}

@inproceedings{lu2025uni,
  title={Uni-layout: Integrating human feedback in unified layout generation and evaluation},
  author={Lu, Shuo and Chen, Yanyin and Feng, Wei and Fan, Jiahao and Li, Fengheng and Zhang, Zheng and Lv, Jingjing and Shen, Junjie and Law, Ching and Liang, Jian},
  booktitle={Proceedings of the 33rd ACM International Conference on Multimedia},
  pages={7709--7718},
  year={2025}
}

@inproceedings{bi2025customttt,
  title={Customttt: Motion and appearance customized video generation via test-time training},
  author={Bi, Xiuli and Lu, Jian and Liu, Bo and Cun, Xiaodong and Zhang, Yong and Li, Weisheng and Xiao, Bin},
  booktitle={Proceedings of the AAAI Conference on Artificial Intelligence},
  volume={39},
  number={2},
  pages={1871--1879},
  year={2025}
}

@article{che2024gamegen,
  title={Gamegen-x: Interactive open-world game video generation},
  author={Che, Haoxuan and He, Xuanhua and Liu, Quande and Jin, Cheng and Chen, Hao},
  journal={arXiv preprint arXiv:2411.00769},
  year={2024}
}

@article{zhang2025u,
  title={U-StyDiT: Ultra-high quality artistic style transfer using diffusion transformers},
  author={Zhang, Zhanjie and Ma, Ao and Cao, Ke and Wang, Jing and Liu, Shanyuan and Ma, Yuhang and Cheng, Bo and Leng, Dawei and Yin, Yuhui},
  journal={arXiv preprint arXiv:2503.08157},
  year={2025}
}

@inproceedings{bi2024using,
  title={Using My Artistic Style? You Must Obtain My Authorization},
  author={Bi, Xiuli and Liu, Haowei and Li, Weisheng and Liu, Bo and Xiao, Bin},
  booktitle={European Conference on Computer Vision},
  pages={305--321},
  year={2024},
  organization={Springer}
}

\clearpage

\appendix

\maketitlesupplementary

\section{Proof to Theorem}
\subsection{Proof of Theorem 4}
For token, ${{t}_{j}}=({{t}_{hj}},{{t}_{wj}})$, each remaining token in the input sequence has an equal probability of being shuffled into the same group as ${{t}_{j}}$. Under this condition, the expected interactive distance between tokens can be calculated as follows:
\begin{equation}
\begin{split}
  d({{t}_{j}}) &= \underset{\left. {{t}_{k}} \right|({{t}_{j}},{{t}_{k}})\in {{I}_{S}}}{\mathop{\mathbb{E}}}\,\left[ {{d}_{S}}({{t}_{j}},{{t}_{k}}) \right] \\
  &= \frac{1}{HW{{d}_{i}}-1} \sum\limits_{h=0}^{H-1} \sum\limits_{w=0}^{W{{d}_{i}}-1} \sqrt{{{(h-{{t}_{hj}})}^{2}}+{{(w-{{t}_{wj}})}^{2}}}
\end{split}
\end{equation}
Using the following mean inequality
\begin{equation}
\sqrt{\frac{{{x}^{2}}+{{y}^{2}}}{2}}\ge \frac{x+y}{2}
\end{equation}
we can calculate a lower bound on $d({{t}_{j}})$:
\begin{equation}
\begin{split}
  d({{t}_{j}}) &=\frac{1}{HW{{d}_{i}}-1}\sum\limits_{h=0}^{H-1}{\sum\limits_{w=0}^{W{{d}_{i}}-1}{\sqrt{{{(h-{{t}_{hj}})}^{2}}+{{(w-{{t}_{wj}})}^{2}}}}} \\
  &\ge \frac{1}{HW{{d}_{i}}-1}\sum\limits_{h=0}^{H-1}{\sum\limits_{w=0}^{W{{d}_{i}}-1}{\frac{\sqrt{2}}{2}(\left| h-{{t}_{hj}} \right|+\left| w-{{t}_{wj}} \right|)}}
\end{split}
\end{equation}
Due to the following formulas:
\begin{equation}
\begin{split}
  \sum\limits_{h=0}^{H-1}{\left| h-{{t}_{hj}} \right|}
  &=\sum\limits_{h=0}^{{{t}_{hj}}}{\left| {{t}_{hj}}-h \right|}+\sum\limits_{h={{t}_{hj}}}^{H-1}{\left| h-{{t}_{hj}} \right|} \\ 
  & =\frac{{{t}_{hj}}({{t}_{hj}}+1)}{2}+\frac{(H-{{t}_{hj}})(H-{{t}_{hj}}-1)}{2}
\end{split}
\end{equation}
and
\begin{equation}
\begin{split}
  \sum\limits_{w=0}^{W{{d}_{i}}-1}{\left| w-{{t}_{wj}} \right|}
  &=\sum\limits_{w=0}^{{{t}_{hj}}}{\left| {{t}_{wj}}-w \right|}+\sum\limits_{w={{t}_{hj}}}^{W{{d}_{i}}-1}{\left| w-{{t}_{wj}} \right|} \\ 
  &=\frac{{{t}_{wj}}({{t}_{wj}}+1)}{2}  \\ 
  &+\frac{(W{{d}_{i}}-{{t}_{wj}})(W{{d}_{i}}-{{t}_{wj}}-1)}{2}
\end{split}
\end{equation}
We can obtain 
\begin{equation}
\begin{split}
  d({{t}_{j}})
  &\ge \frac{\sqrt{2}}{4(HW{{d}_{i}}-1)}[H{{m}_{wj}}({{m}_{wj}}+1)\\
  &+W{{d}_{i}}{{m}_{hj}}({{m}_{hj}}+1)\\ 
  &+H(W{{d}_{i}}-{{m}_{wj}})(W{{d}_{i}}-{{m}_{wj}}-1)\\
  &+W{{d}_{i}}(H-{{m}_{hj}})(H-{{m}_{hj}}-1)]
\end{split}
\end{equation}
Thus proving the original theorem.
\subsection{Proof of Corollary 5}
For tokens within the same local group, we calculate their average interactive distance:
\begin{equation}
\bar{d}({{t}_{j}})=\frac{1}{{{s}^{2}}-1}\sum\limits_{j=1}^{{{s}^{2}}-1}{d({{t}_{j}}\left| HW{{d}_{i}}-j \right.)}
\end{equation}
Among these, $d({{t}_{j}}\left| HW{{d}_{i}}-j \right.)$ represents the expected interactive distance when the number of remaining tokens is $HW{{d}_{i}}-j$. Based on Assumption~\ref{def:Local 3D Partition}, we know $j\in [2,{{s}^{2}}-1]$, and we can derive the following approximation:
\begin{equation}
d({{t}_{j}}\left| HW{{d}_{i}}-j \right.)\approx d({{t}_{j}}\left| HW{{d}_{i}}-1 \right.)=d({{t}_{j}})
\end{equation}
Therefore, the value of $\bar{d}({{t}_{j}})$ can be estimated as follows:
\begin{equation}
\begin{split}
    \bar{d}({{t}_{j}})
  & = \frac{1}{{{s}^{2}}-1} \sum\limits_{j=1}^{{{s}^{2}}-1} d({{t}_{j}}\left| HW{{d}_{i}}-j \right.) \\ 
  & \approx \frac{1}{{{s}^{2}}-1} \sum\limits_{j=1}^{{{s}^{2}}-1} d({{t}_{j}}\left| HW{{d}_{i}}-1 \right.) \\
  & = \frac{1}{{{s}^{2}}-1} \sum\limits_{j=1}^{{{s}^{2}}-1} d({{t}_{j}}) \\
  & = d({{t}_{j}})
\end{split}
\end{equation}

\section{Training Details}
\label{paper: Training Details}
\textbf{Training and Testing Datasets.} We curated a dataset containing 1.73 million high-resolution, high-quality images, each with an aesthetic score of 5.5 or higher. Following the production methodology outlined in ControlNet \cite{zhang2023controlnet}, we generated various types of conditional images, including HED, Canny, Depth, and Segment. For the quantitative experiments, the model was trained at a resolution of 512 and evaluated on the COCO validation dataset \cite{lin2014microsoftcoco}, consisting of 5000 images. For qualitative visual evaluation, the model was trained at a resolution of 1024 and tested on a high-quality, high-resolution test set of 1000 images. To ensure a fair evaluation, all experiments were conducted with a training of 5 epochs on 16 NVIDIA A100 GPUs, using identical settings across all trials.

\subsection{Additional experimental results}
\subsubsection{More exploration on Relevance}
\begin{figure}[!htb]
\centering
\includegraphics[width=\linewidth]{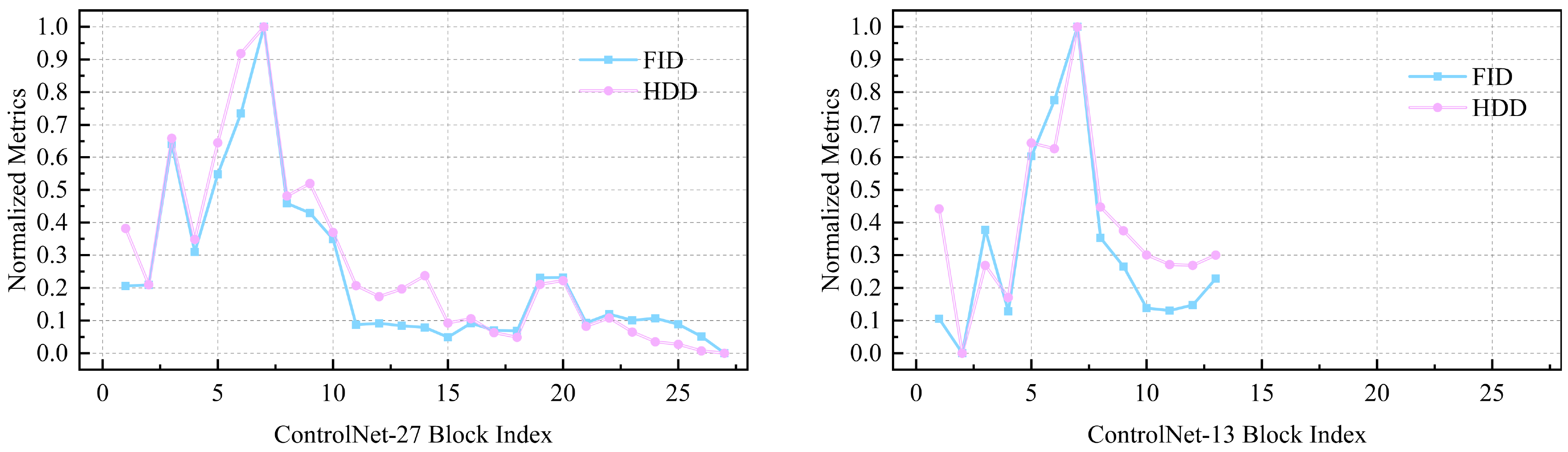}
\caption{\label{img:appendix_full1}Impact of deleting specific locations on the generated metrics in ControlNet with 27 and 13 blocks.
}
\end{figure}
In the main text, we conducted a study of DiT-ControlNet Relevance based on the PixArt-$\delta$ with 27 control blocks and obtained the following insights: the most critical modules of DiT-ControlNet are concentrated in the early-middle layers (e.g., blocks 5, 6, and 7). Overall, the ControlNet Relevance Score exhibits a trend of initially increasing and then decreasing.

To further validate the generalizability of this observation, we also conducted a similar study on a PixArt model with only the first 13 control blocks. Fig.~\ref{img:appendix_full1} presents the normalized FID and HDD metrics of the generated images after removing the control module at a specific position. It is evident that the experimental results align with our previous findings: the most influential blocks in the control branch remain blocks 5, 6, and 7, which follow the same trend of increasing and then decreasing in relevance. This demonstrates the robustness of our observations, which can provide valuable guidance for subsequent specific design choices.

\begin{figure}[!htb]
\centering
\includegraphics[width=0.9\linewidth]{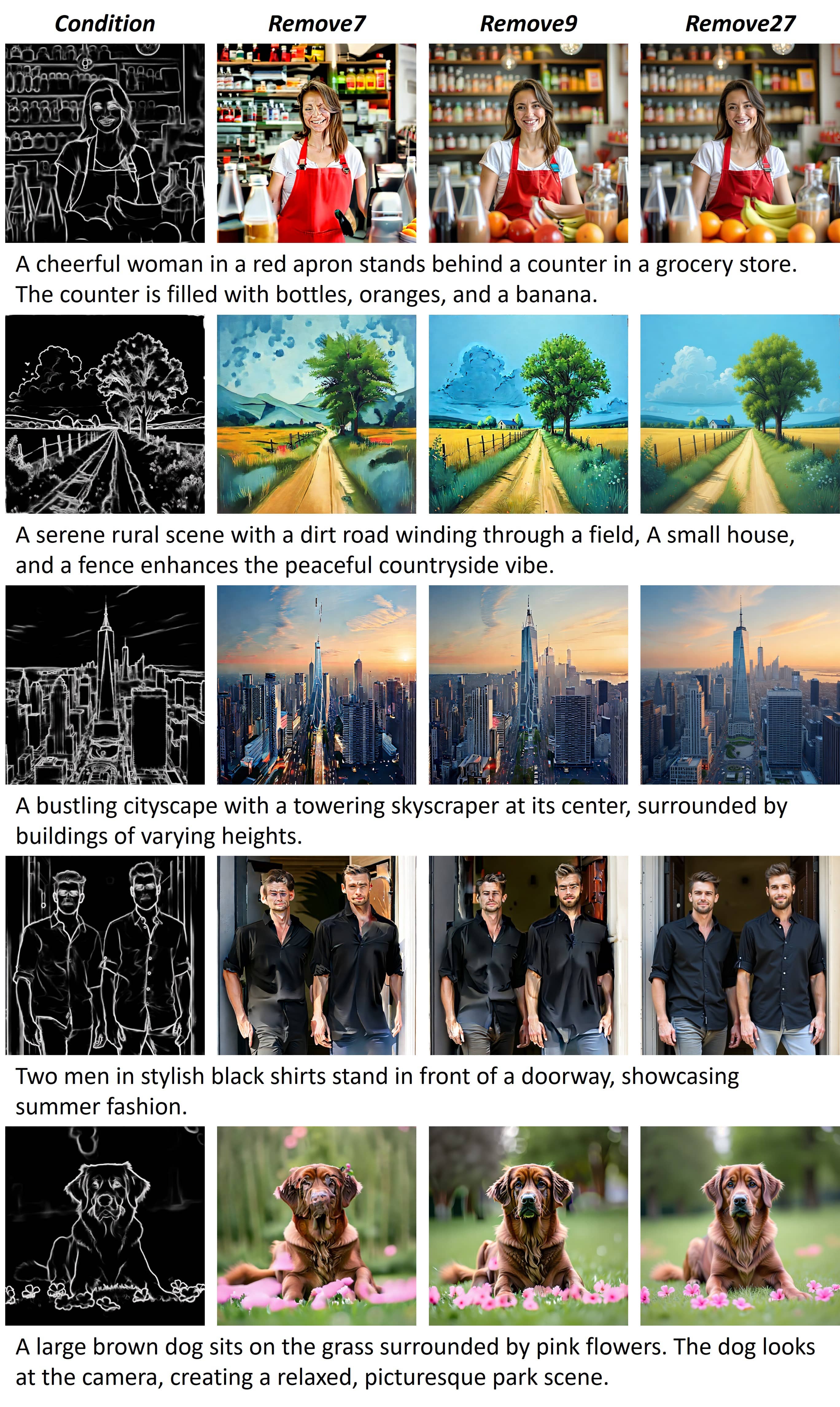}
\caption{\label{img:append_remove}Additional visual results of skipping specific ControlNet layers (7, 9, and 27), correspond to the highest, moderate, and lowest impact on the generated image.
}
\end{figure}

In Fig.~\ref{img:append_remove}, we present additional visual demonstrations of skipping specific ControlNet layers, specifically layers 7, 9, and 27, which correspond to the highest, moderate, and lowest impact on the generated image. The results illustrate that removing layer 7 significantly degrades both the quality of the generated image and the control accuracy. In contrast, skipping the later layers, such as layer 27, has minimal negative effects on the overall performance.

\subsubsection{More ablation results}
To further investigate the effectiveness of the proposed modules, we conduct additional ablation studies focusing on two aspects: the performance of the TDSM and the impact of randomly selecting control positions. The detailed results are summarized in Table~\ref{tab: more abl}.

Table~\ref{tab: more abl}(a) illustrates that the random shuffle mechanism embedded in TDSM significantly enhances the interaction range within the model, facilitating more efficient extraction and propagation of control signals. Notably, replacing TDSM with conventional self-attention not only increases computational overhead but also degrades the generative performance. Furthermore, since the RGLC block simplifies the original PixArt Transformer block, we compare Setting 3, which retains a full feed-forward network and cross-attention, with the final RelaCtrl configuration. Despite a substantial increase in both parameter count and computational complexity, the fully structured Transformer block still underperforms compared to RelaCtrl, highlighting the superior efficiency and design of TDSM.

In addition, we include ablation results where 13 control blocks are randomly selected, as shown in Table~\ref{tab: more abl}(b). Compared to the ControlNet-top13 and Relevance-top13 settings, random selection results in noticeable degradation in both control accuracy and visual fidelity. This observation supports our previous findings: control relevance is more prominent in the early and middle layers of the DiT backbone, whereas random selection tends to distribute control modules more uniformly across the network (as illustrated in Fig. 1 of the manuscript). Because modules placed in deeper layers are less effective at integrating control signals, their contribution to overall performance is diminished.
\begin{table}[!htb]
    \centering
    \small
    \renewcommand{\arraystretch}{1.2}
    \renewcommand{\tabcolsep}{3pt}
    \caption{Supplementary ablation study results.}
    \label{tab: more abl}
    \begin{minipage}{0.48\textwidth}
        \centering
        \textbf{(a) TDSM Ablation Results} \\
        \vspace{2pt}
        \begin{tabular}{cccc}
            \hline
            Setting & HDD$\downarrow$ & FID$\downarrow$ & Para Ratio \\
            \hline
            w/o Shuffle      & 97.29  & 21.63    & 15.70\% \\  
            TDSM to SA       & 96.37  & 21.32    & 30.70\%  \\ 
            RGLC to Copy     & 95.57  & 21.28    & 84.6\%  \\ 
            \rowcolor{gray!20}
            RelaCtrl         & 94.04  & 20.34    & 15.3\%  \\ 
            \hline
        \end{tabular}
    \end{minipage}
    \hfill
    \begin{minipage}{0.48\textwidth}
        \centering
        \textbf{(b) Random Control Module Placement} \\
        \vspace{2pt}
        \begin{tabular}{cccc}
            \hline
            Setting & HDD$\downarrow$ & FID$\downarrow$ & Para Ratio \\
            \hline
            Random-13         & 96.51  & 23.74   & 100\%  \\ 
            ControlNet-top13  & 96.26  & 21.38   & 100\% \\ 
            \rowcolor{gray!20}
            Relevance-top13   & 94.57  & 20.31   & 100\%  \\ 
            \hline
        \end{tabular}
    \end{minipage}
\end{table}

\subsubsection{Experiment on Flux}
\label{appendix:flux_explore}
In this section, we apply RelaCtrl to the latest Flux.1-dev with 12 billion parameters for the experiment. The comparison methods include OminiControl \cite{tan2024ominicontrol} and Flux ControlNet, which is implemented based on the Diffusers library. Flux ControlNet utilizes a standard configuration consisting of a double-stream layer with the first four replication layers, alongside a single-stream layer with the first ten replication layers.

Our goal in integrating RelaCtrl into Flux is to transfer as much of the original architectural insight and design as possible, thereby evaluating the generalization ability of the proposed method. To this end, we proportionally scale the control positions from PixArt to match the architectural layout of Flux. Notably, in the double-stream layers, the control is applied only to the image branch; in the single-stream layers, the control signals are injected exclusively into the image tokens.

Table~\ref{tab:flux1} presents the performance and efficiency comparison of various methods. GFLOPs are calculated at an image resolution of 512, while inference time is measured using 30 DDIM sampling steps at the same resolution. RelaCtrl demonstrates superior performance across multiple quality metrics related to image generation. Although OminiControl introduces only a modest increase in parameter count, it results in a substantial rise in computational cost and inference time, thereby severely constraining its practical applicability. Flux-ControlNet exhibits moderate overhead across various efficiency indicators.

\begin{figure}[!htb]
\centering
\includegraphics[width=\linewidth]{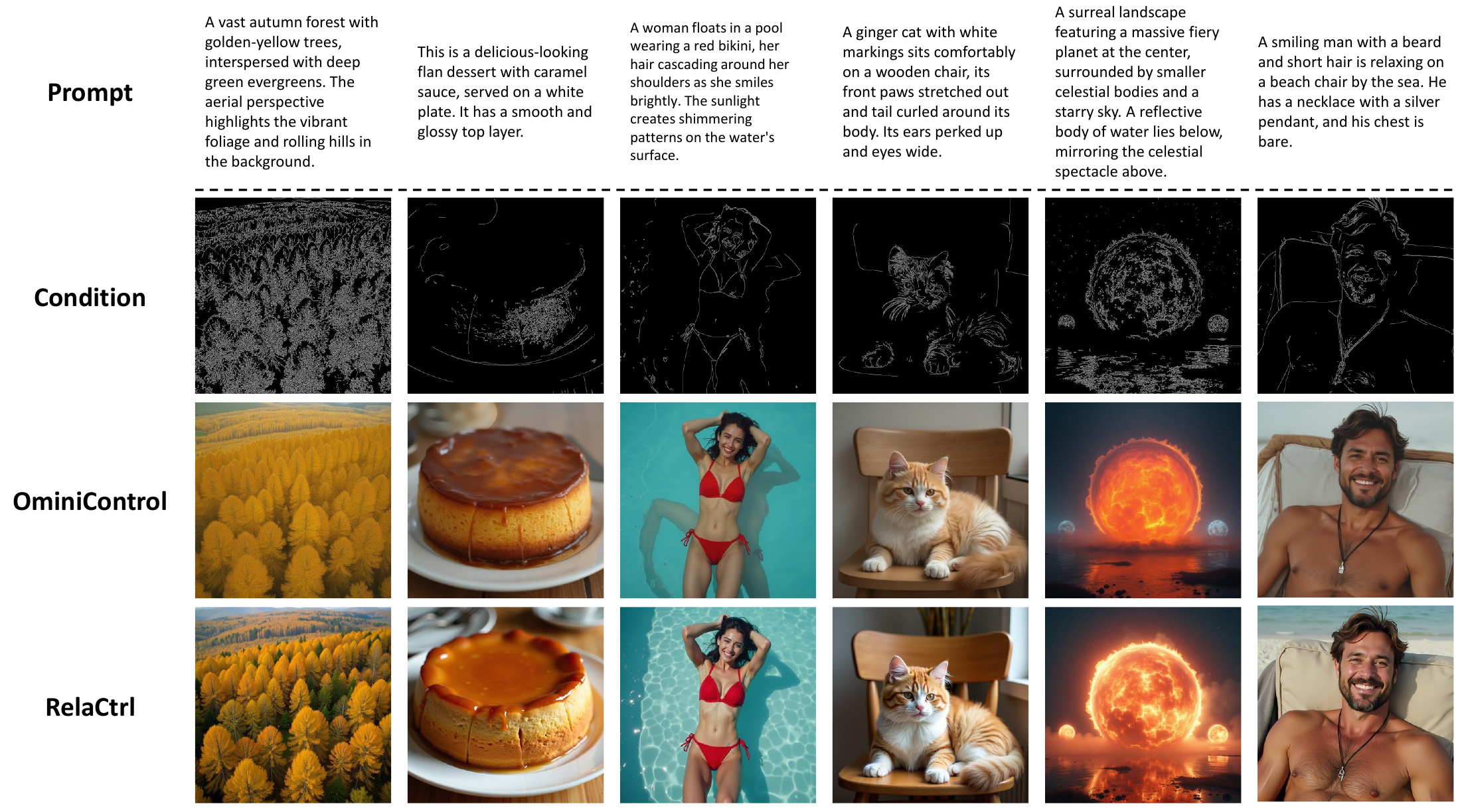}
\caption{\label{img:append_flux}Visual comparison of RelaCtrl and OminiControl on Flux.1-dev.
}
\end{figure}

In contrast, RelaCtrl for Flux achieves a notable reduction in computational complexity and inference latency, striking a favorable balance between efficiency and model size. As illustrated in Fig.~\ref{img:append_flux}, RelaCtrl yields visually precise and detailed control outcomes, while also offering significant computational advantages. These results collectively demonstrate the effectiveness, efficiency, and generalizability of RelaCtrl across different DiT-based architectures.
\begin{table*}[]
    \centering
    \small
    \renewcommand{\arraystretch}{1.4}
\renewcommand{\tabcolsep}{3pt}
\caption{Quantitative comparison results based on the Flux.1-dev. The best results are in bold, and the second-best results are underlined.}
    \label{tab:flux1}
\begin{tabular}{c|c|c|c|c|c|c|c}
\hline Setting & FID $\downarrow$ & HDD $\downarrow$ & \begin{tabular}{c} 
CLIP- \\
Score $\uparrow$
\end{tabular} & \begin{tabular}{c} 
CLIP- \\
Ae $\uparrow$
\end{tabular} & Parameters (M) & \begin{tabular}{c} 
Complexity \\
(GFLOPs)
\end{tabular} & \begin{tabular}{c} 
Inference \\
(s)
\end{tabular} \\
\hline Flux.1-dev & - & - & - & - & 11901.39 & 9925.78 & 4.78 \\
\hline +ControlNet-canny & 17.61 & \underline{87.26} & 0.2813 & \textbf{5.2871} & +2952.65  &\underline{+2578.33} & \underline{+0.79} \\
\hline +OminiControl-canny & \underline{17.19} & 87.38 & \underline{0.2837} & 5.1591 & \textbf{+14.49} & +6637.41 & +3.11 \\
            \rowcolor{gray!20}
\hline +RelaCtrl-canny & \textbf{16.86} & \textbf{86.87} & \textbf{0.2861} & \underline{5.2039} & \underline{+549.19} & \textbf{+495.03} & \textbf{+0.34} \\
\hline
\end{tabular}
\end{table*}

\subsubsection{Inference with Community Models}
\begin{figure}[!htb]
\centering
\includegraphics[width=\linewidth]{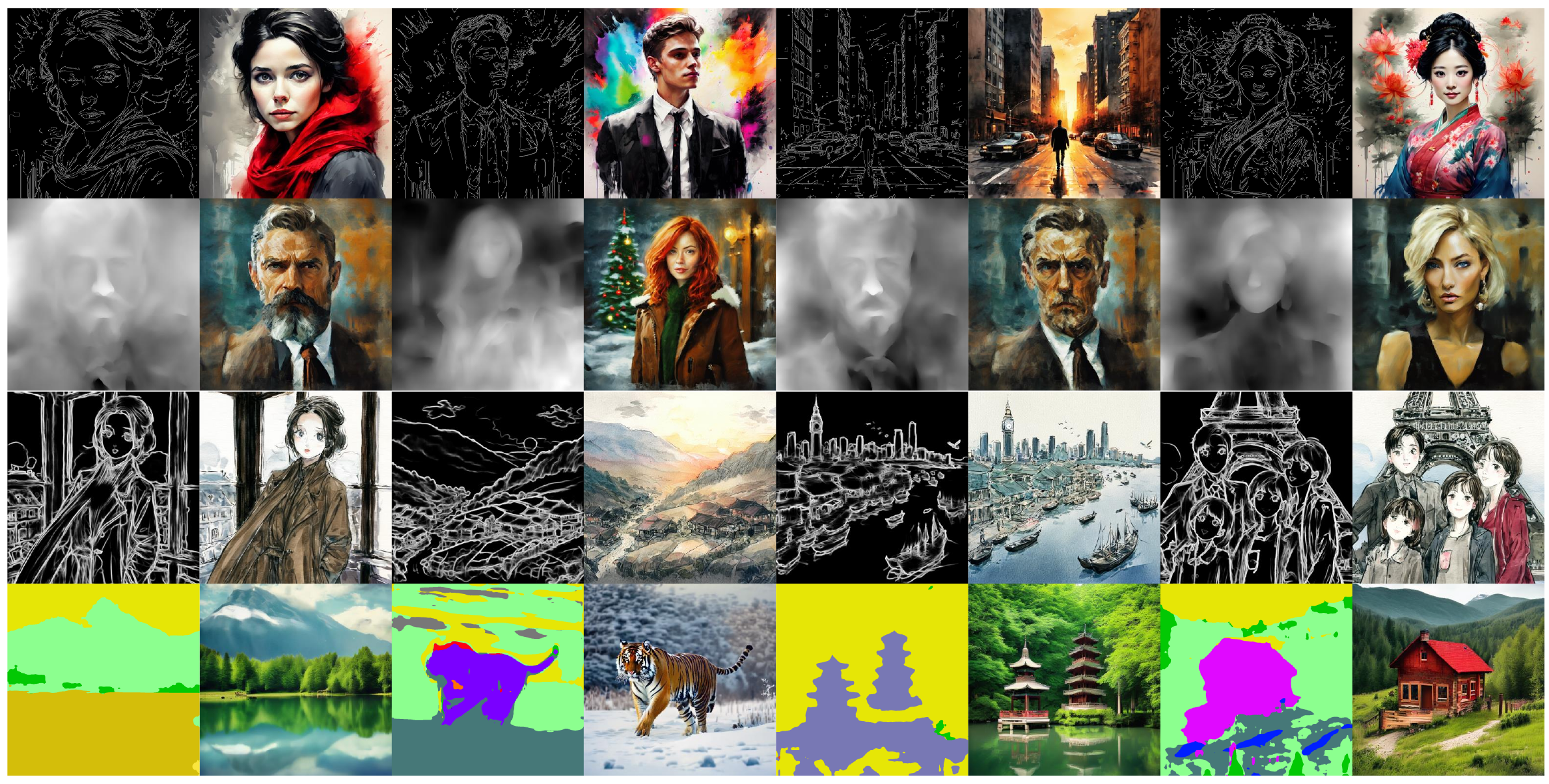}
\caption{\label{img:append_style}The control effect of RelaCtrl on the fine-tuned PixArt model. The upper and lower rows show the four transitions: (1) Canny to paint, (2) Depth to oil, (3) HED to gufeng, and (4) Segmentation to pixel.
}
\end{figure}

We perform inference using the PixArt weights, which were fine-tuned with Lora. Although RelaCtrl has not previously been trained in these weights, it can still utilize them effectively. Fig.~\ref{img:append_style} showcases the model's generated paint, oil, gufeng, and pixel-style images under the specified conditions.

\subsubsection{Efficiency across diverse platforms}
To further evaluate the computational efficiency of RelaCtrl across diverse hardware platforms, we conduct comprehensive runtime benchmarks on three distinct devices: NVIDIA A100, NVIDIA RTX 3090, and Huawei Ascend-910B. The RTX 3090 represents a widely adopted consumer-grade GPU, whereas the A100 and Ascend-910B are high-performance, industrial-grade accelerators featuring fundamentally different architectures. Detailed benchmark results are reported in Table~\ref {tab:full_efficiency}. Across all tested platforms and workloads, our method consistently demonstrates high runtime efficiency and stable performance. These results further underscore the robustness and hardware-agnostic efficiency of RelaCtrl, highlighting its practicality for deployment in both consumer-level and industrial-scale applications.

\begin{table*}[htp]
    \centering
    \small
    \renewcommand{\arraystretch}{1.2}
\renewcommand{\tabcolsep}{3pt}
\caption{Effectiveness evaluation across different hardware platforms.}
\label{tab:full_efficiency}
\begin{tabular}{c|c|c|c|c|c}
\hline Methods & Base model & Parameters & Complexity & \begin{tabular}{l}
Inference
(20 steps)
\end{tabular} & Memory \\
\hline Pixart-a & 

& 611.15 M & \begin{tabular}{l}
542.56 on A100 \\
266.78 on 3090 \\
479.01 on 910B2
\end{tabular} & \begin{tabular}{l}
3.81 Sec on A100 \\
9.42 Sec on 3090 \\
5.92 Sec on 910B2
\end{tabular} & \begin{tabular}{l}
2114 MiB on A100 \\
2130 MiB on 3090 \\
2297 MiB on 910B2
\end{tabular} \\
\cline{1-1} \cline{3-6} +ControlNet & \text { Pixart-a }  & +294.34 M & $
\begin{tabular}{l}
+270.57 on A100 \\
+128.61 on  3090 \\
+246.16 on  910B2
\end{tabular}
$ & \begin{tabular}{l}
+0.51 Sec on A100 \\
+0.59 Sec on 3090 \\
+3.99 Sec on 910B2
\end{tabular} & $
\begin{tabular}{l}
+1694 MiB on A100\\
+1743 MiB on 3090\\
+1722 MiB  on 910B2
\end{tabular}
$ \\
\cline{1-1} \cline{3-6} +RelaCtrl & & +45.15M & \begin{tabular}{l}
+46.71 on A100 \\
+28.50 on 3090 \\
+40.65 on 910B2
\end{tabular} & \begin{tabular}{l}
+0.24 Sec on A100 \\
+0.44 Sec on 3090 \\
+1.24 Sec on 910B2
\end{tabular} & \begin{tabular}{l}
+395 MiB on A100 \\
+582 MiB on 3090 \\
+735 MiB on 910B2
\end{tabular} \\
\hline
\end{tabular}
\end{table*}

\end{document}